\documentclass[11pt]{article}
\usepackage[left=25mm, right=25mm, top=25mm, bottom=25mm, includehead=true, includefoot=true]{geometry}

\usepackage[authoryear,round]{natbib}

\usepackage{graphicx}
\usepackage{url}
\usepackage{authblk} 
\usepackage[parfill]{parskip} 

\usepackage{sectsty}
\sectionfont{\normalsize}
\subsectionfont{\normalsize}
\subsubsectionfont{\normalsize}
\paragraphfont{\normalsize}

\usepackage[pdftex]{hyperref} 
\hypersetup{pdfborder={0 0 0} } 

\usepackage{color}
\usepackage{fancyvrb}

\DefineVerbatimEnvironment{Highlighting}{Verbatim}{commandchars=\\\{\}}
\usepackage{framed}
\definecolor{shadecolor}{RGB}{248,248,248}

\usepackage{url}
\usepackage{boxedminipage}
\usepackage{textpos}

\title{Urban Comfort Assessment in the Era of Digital Planning: A Multidimensional, Data-driven, and AI-assisted Framework}

\author[1,2]{Sijie Yang}
\author[1]{Binyu Lei}
\author[1,3]{Filip Biljecki\thanks{filip@nus.edu.sg}}
\affil[1]{Department of Architecture, National University of Singapore}
\affil[2]{School of Engineering and Applied Science, University of Pennsylvania}
\affil[3]{Department of Real Estate, National University of Singapore}

\date{}

\begin{document}

\maketitle


\begin{abstract}

\begin{textblock*}{\textwidth}(0cm,-10cm)
\begin{center}
\begin{footnotesize}
\begin{boxedminipage}{1\textwidth}
This is the Accepted Manuscript version of a conference paper presented at 19th International Conference on Computational Urban Planning and Urban Management (CUPUM 2025) on Jun 24, 2025.\\ Cite as:
Yang S, Lei B, Biljecki F (2025): Urban Comfort Assessment in the Era of Digital Planning: A Multidimensional, Data-driven, and AI-assisted Framework. In: \textit{19th International Conference on Computational Urban Planning and Urban Management (CUPUM 2025)}, London, UK.
\end{boxedminipage}
\end{footnotesize}
\end{center}
\end{textblock*}

\centering

Ensuring liveability and comfort is one of the fundamental objectives of urban planning. Numerous studies have employed computational methods to assess and quantify factors related to urban comfort such as greenery coverage, thermal comfort, and walkability. However, a clear definition of urban comfort and its comprehensive evaluation framework remain elusive. Our research explores the theoretical interpretations and methodologies for assessing urban comfort within digital planning, emphasising three key dimensions: multidimensional analysis, data support, and AI assistance.


\end{abstract}


\section{The Multidimensional Nature of Urban Comfort}\label{urban comfort concept}

Comfort is defined as the degree of satisfaction with the environment of a person \citep{frontczak2012quantitative, shin2016toward}. In the built environment, there are many environmental factors that have been proven to be associated with human comfort sensation, including visual elements \citep{klemm2015street, liu2023towards, yang2025thermal}, physiological data of individuals \citep{chwalek2024dataset}, occupant behaviour \citep{miller2025make}, meteorological indicators \citep{xie2019outdoor, miller2023introducing, mosteiro2024people}, the design of public places \citep{santos2018approaches}, the dynamics of life quality \citep{lei2025developing} and so on. Urban comfort, in this context, refers specifically to the experience of comfort within the urban environment. It is inherently a multidimensional concept due to the complexity and variability of the urban physical environment. Unlike general comfort, which can be defined at the individual level, urban comfort emerges as a collective and spatially embedded phenomenon, shaped by the interplay between environmental attributes and human perception. The diverse components of the urban environment evoke a wide range of perceptions in individuals \citep{salesses2013collaborative, dubey2016deep, qiu2023subjective, yang2023role, ito2024understanding}, making it unsuitable for reduction into a single numerical index or one-dimensional measure, as it is often the case with other concepts in planning such as walkability. In addition, comfort is highly subjective, varying significantly from person to person based on their preferences, experiences, and expectations \citep{castaldo2018subjective, wang2018uncertainty}, which adds another layer of complexity to the understanding of urban comfort.

    \begin{figure}[ht]
        \centering
        \includegraphics[width=0.9\linewidth]{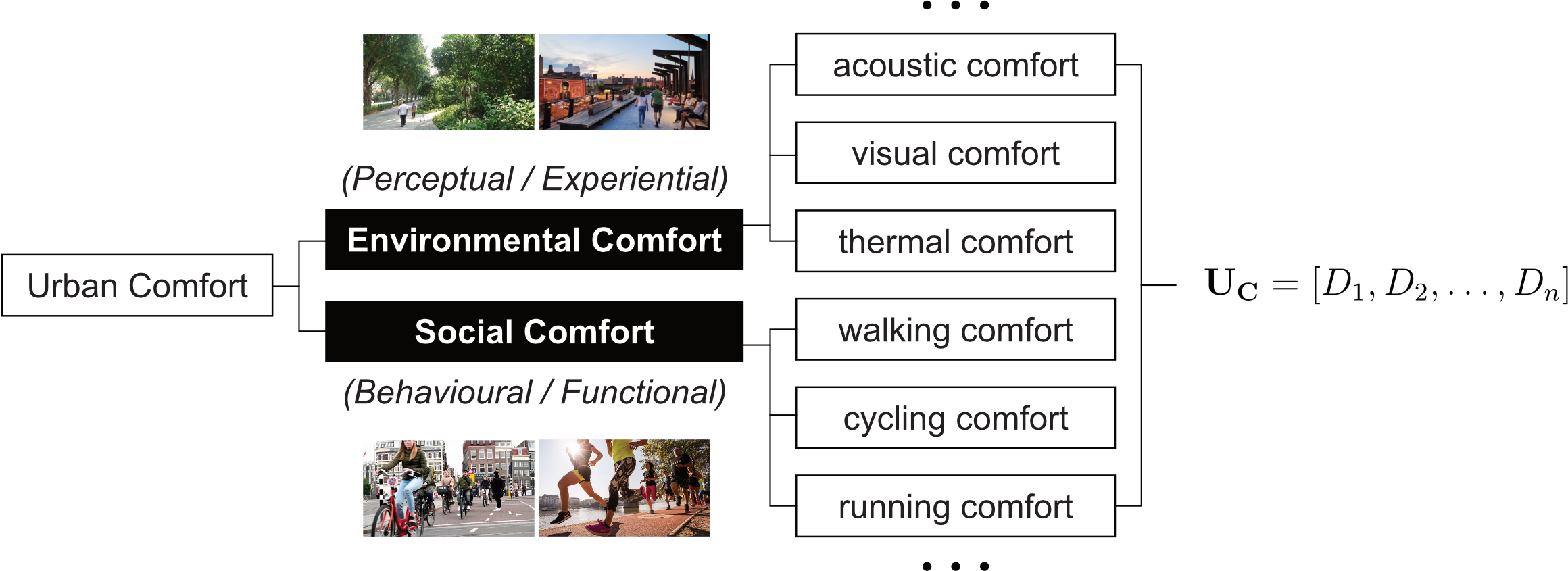}
        \caption{Proposed multidimensional framework of urban comfort.}
        \textbf{\label{fig:framework_concept}}
    \end{figure}

Broadly, urban comfort can be categorised into two main dimensions: environmental comfort and social comfort, as described in \textbf{Figure} \ref{fig:framework_concept}. Environmental comfort primarily refers to the perceptual and experiential satisfaction of people with the physical aspects of the built environment, such as thermal comfort \citep{rupp2015review}, visual comfort \citep{lam2020cross}, and acoustic comfort \citep{yang2005acoustic}. In contrast, social comfort relates to how well the urban environment supports human interactions and behavioural needs. It emphasizes the functional ability of spaces to facilitate desired user activities, such as walkability, accessibility, and safety. Examples include walking comfort \citep{ma2021critical} and cycling comfort \citep{ayachi2015identifying}, which reflect how urban infrastructure influences ease of movement and user experience.

Urban comfort is deeply intertwined with the characteristics of the urban environment \citep{gulyas2006assessment}. People experience different levels of comfort in heterogeneous urban settings \citep{kim2022linking}, and even within the same environment, individuals may perceive comfort differently across various dimensions \citep{du2023multiple}. This dual-layer relationship indicates that urban comfort is not only context-dependent but also multidimensional. Instead of considering urban comfort as a single scalar value, it is more accurate to conceptualise it as a vector $\mathbf{U_C}$ that captures the multidimensional nature of urban comfort across several dimensions. Mathematically, this can be expressed as $\mathbf{U_C} = [D_1, D_2, \ldots, D_n]$, where:

 \begin{itemize}
     \item $D_i$ represents each dimension of urban comfort (e.g., thermal comfort, visual comfort).
     \item $\mathbf{U_C}$ is the vector of all the dimensions of comfort, reflecting the multidimensional aspects of urban comfort without reducing them to a single aggregate value.
 \end{itemize}

Our vector-based approach acknowledges that different dimensions of urban comfort can interact with each other, contributing differently to the perception of comfort depending on the urban context and individual preferences. The complexity of assessing urban comfort lies in understanding how these dimensions coexist and influence each other, rather than simply aggregating them into a single index. This perspective reinforces the need for a comprehensive evaluation framework that captures the nuanced and multidimensional experiences of individuals within urban environments.

\begin{figure}[ht]
        \centering
        \includegraphics[width=0.8\linewidth]{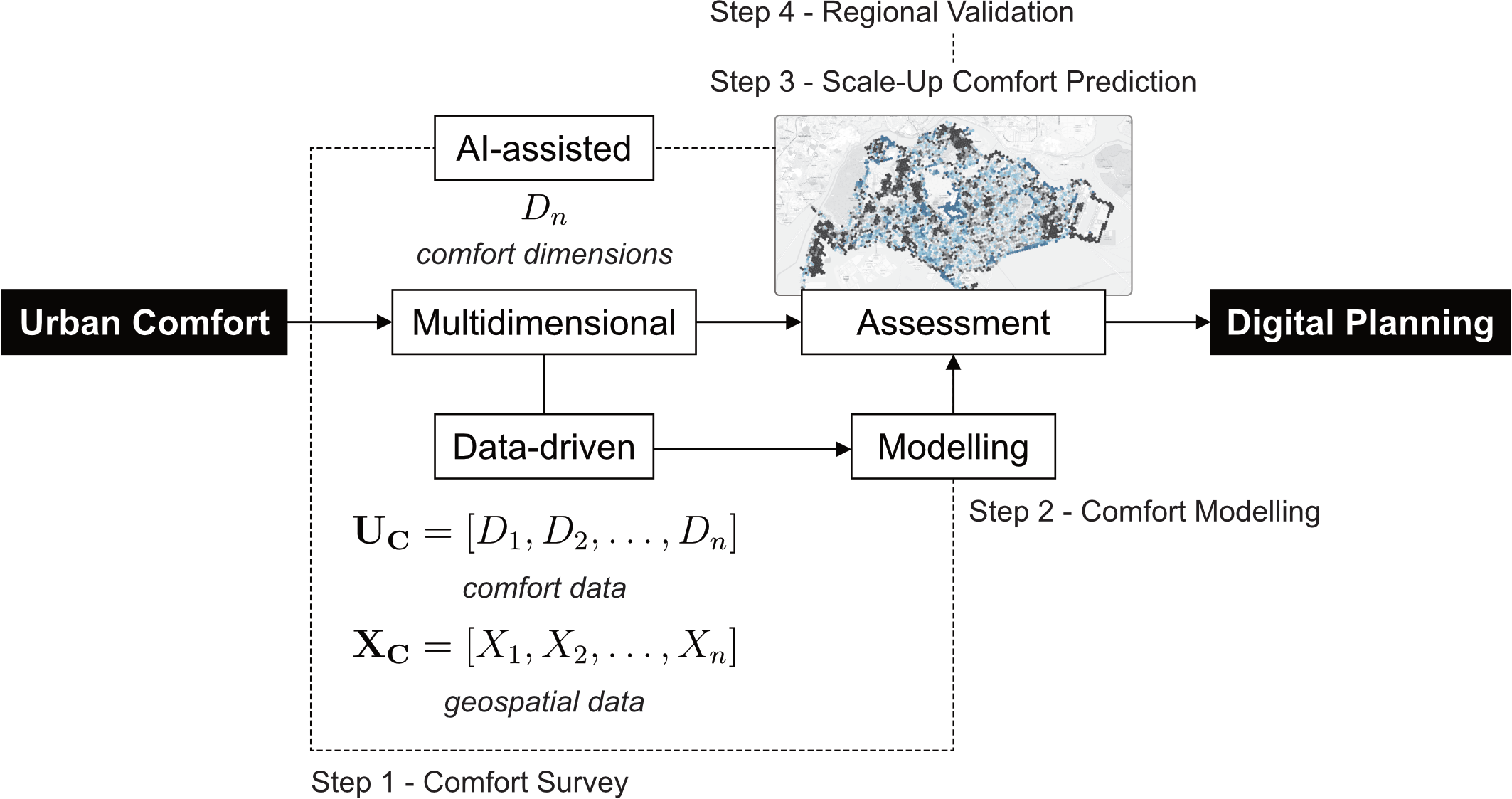}
        \caption{Urban comfort assessment framework in this study.}
        \textbf{\label{fig:framework_overall}}
    \end{figure}

In this study, we clarified the multi-dimensional nature of urban comfort and its math expression by definition. In the following sections, we further propose the four steps of urban comfort assessment and the role of AI algorithms in modelling urban comfort for digital planning, as shown in \textbf{Figure} \ref{fig:framework_overall}.

\section{Four-Step Framework for Data-Driven Urban Comfort Assessment}\label{urban comfort assessment framework}

    \begin{figure}[ht]
        \centering
        \includegraphics[width=0.9\linewidth]{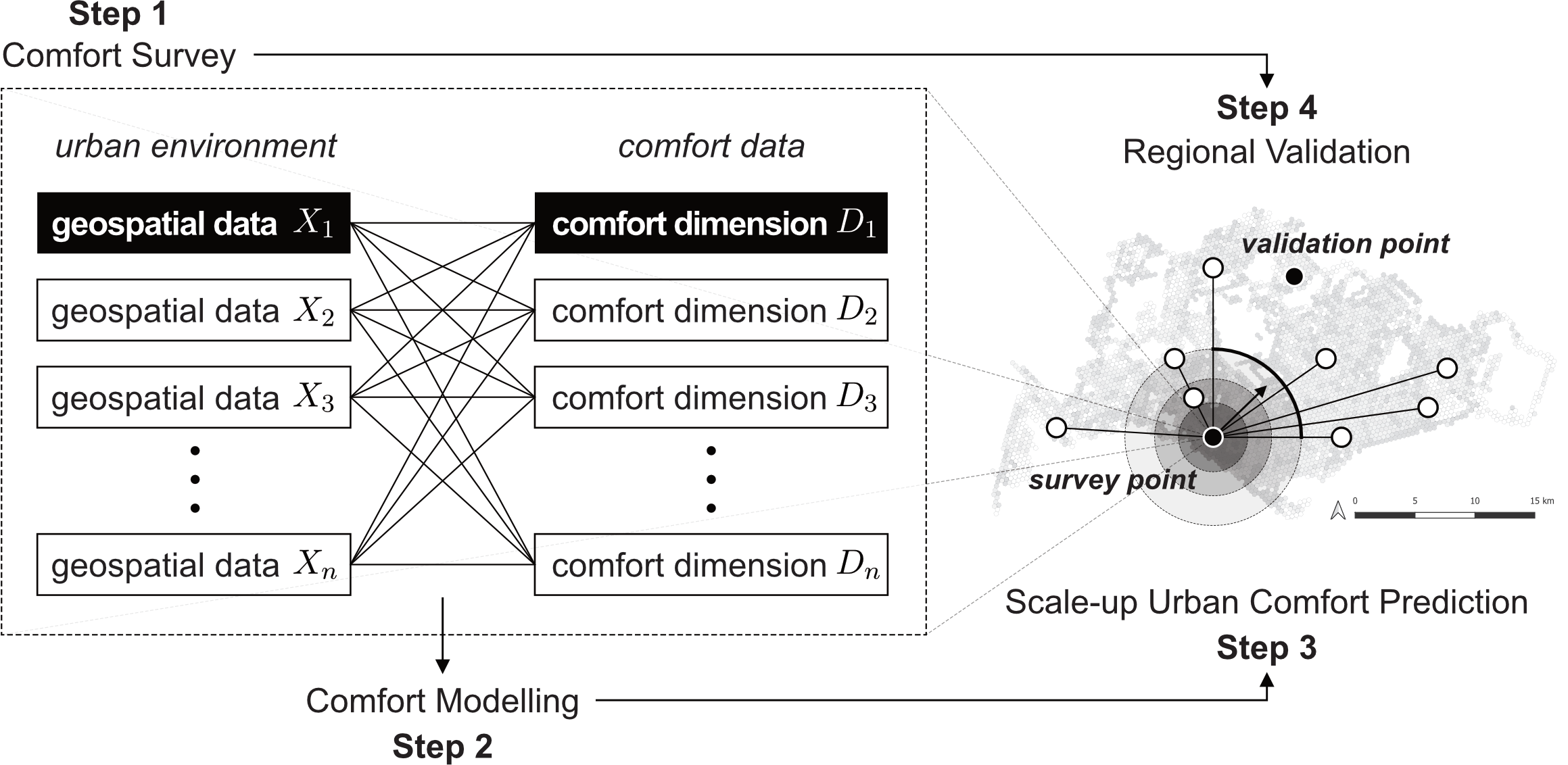}
        \caption{Data-driven urban comfort assessment framework.}
        \textbf{\label{fig:framework_data}}
    \end{figure}

To effectively evaluate urban comfort in its multidimensional attributes, we propose a structured, data-driven assessment framework encompassing four key steps: conducting comfort surveys, developing comfort models, scaling up urban comfort predictions, and performing regional validation, as illustrated in \textbf{Figure} \ref{fig:framework_data}. Each step is data-supported, leveraging three essential data streams: geospatial data to characterise the urban environment in the comfort survey, survey results to capture human comfort perception, and additional comfort survey data from distinct sites for model validation.

As shown in \textbf{Figure} \ref{fig:framework_data}, the process begins with a comfort survey conducted at a specified urban location, recording geospatial characteristics ($X_1\ldots X_n$) and various comfort dimensions ($D_1\ldots D_n$). Subsequently, the collected geospatial and comfort data undergo modelling to capture the non-linear relationship between the urban environment and human comfort. This model enables scaling up predictions to more urban areas across the city based on extended geospatial data, facilitating an urban-scale comfort assessment. Finally, the comfort model is validated using survey data from an additional, designated validation location.

\section{AI-assisted Comfort Modelling in the Era of Digital Planning}\label{ai comfort modelling}

    \begin{figure}[ht]
        \centering
        \includegraphics[width=0.9\linewidth]{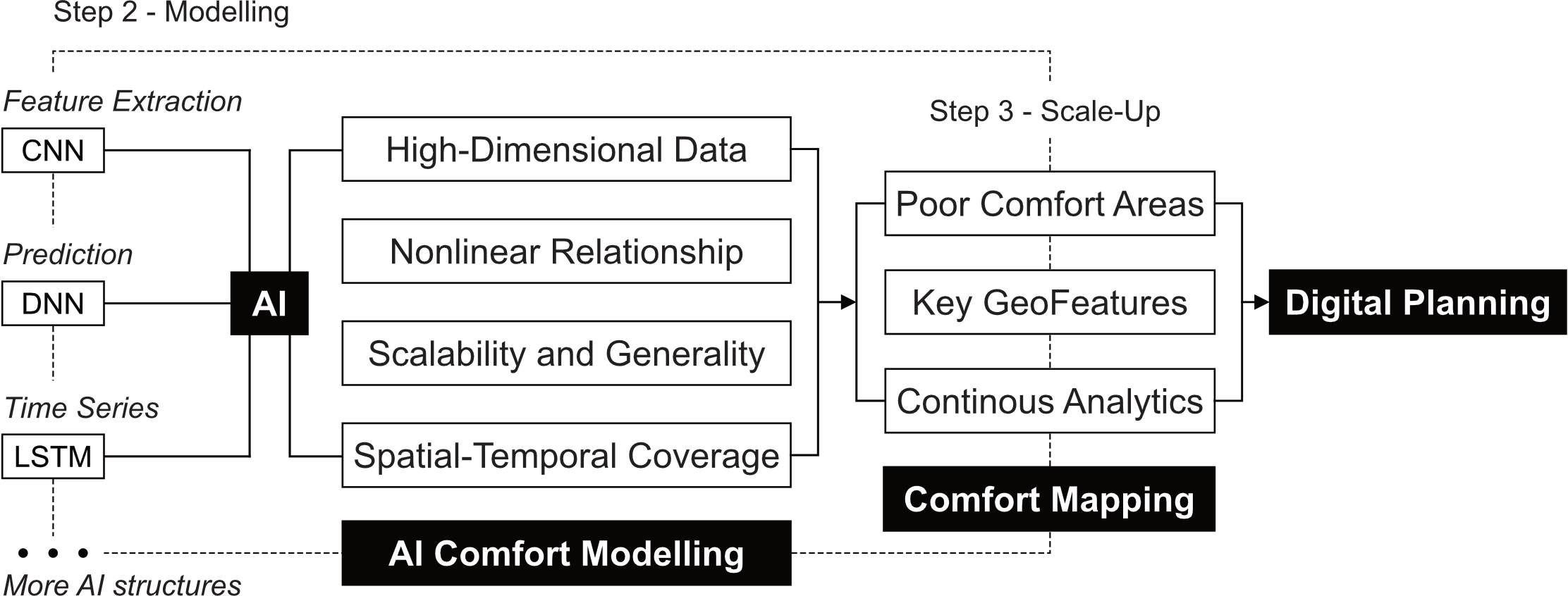}
        \caption{AI-assisted comfort modelling features for digital planning.}
        \textbf{\label{fig:framework_ai}}
    \end{figure}

Data-driven, multidimensional urban comfort modelling can be further enhanced by artificial intelligence to support digital planning more effectively (\textbf{Figure} \ref{fig:framework_ai}). AI models offer significant advantages, including high-dimensional data representation, non-linear relationship modelling, scalability, generalisability, and spatial-temporal understanding, making them better suited than traditional methods for tackling the complexities of comfort modelling. For instance, convolutional neural network (CNN) excels in feature extraction from image data and are widely used in processing satellite and street view imagery (SVI) as data representation tools. Advanced AI models, such as deep neural network (DNN), demonstrate robust predictive performance with non-linear problems like human comfort and possess strong scalability and generalisation, enabling application to untrained data. Comfort models developed for one urban area can be extended to predict comfort in other regions. Additionally, long-short-term memory network (LSTM) is well-suited for time-series data, making them valuable for handling large volumes of temporally-labelled geographic data in dynamic urban observations.

These AI technologies, when integrated, can significantly enhance comfort modelling and mapping within our data-supported urban comfort assessment framework, offering powerful tools for digital urban planning. As depicted in \textbf{Figure} \ref{fig:framework_ai}, typical applications include identifying areas with low comfort for prioritisation in urban renewal, highlighting critical urban environmental elements needing quality improvement, and conducting continuous geospatial analyses based on spatial-temporal analytics to inform urban planning decisions \citep{yap2023incorporating, liang2023revealing, yang2025thermal}.

\section{Case Study: Thermal Comfort Analytics in Singapore}\label{case study}

    \begin{figure}[ht]
        \centering
        \includegraphics[width=0.9\linewidth]{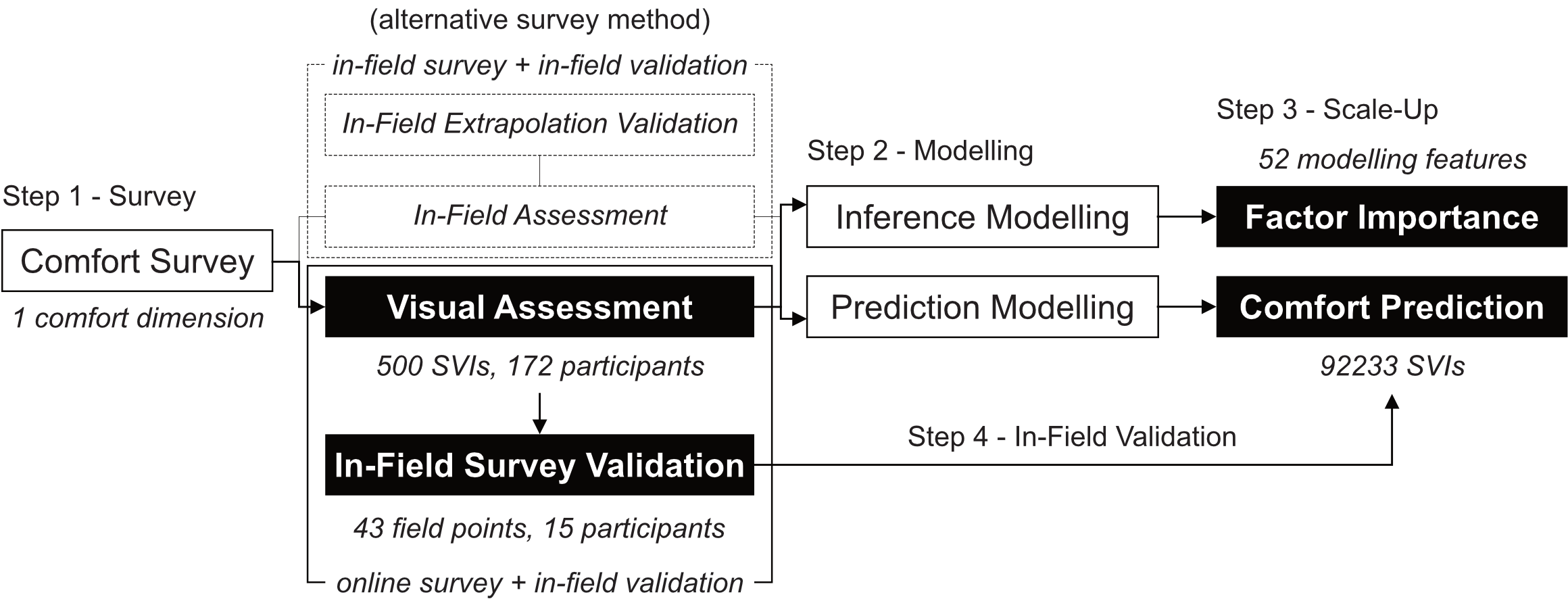}
        \caption{Urban comfort assessment framework in the case study.}
        \textbf{\label{fig:framework_case study}}
    \end{figure}

To implement our proposed four-step, AI-assisted urban comfort assessment framework, we selected Singapore as the study area for a focused analysis of urban thermal comfort, given its strong emphasis on urban liveability and resilience in a high-density, tropical environment. As illustrated in \textbf{Figure} \ref{fig:framework_case study}, this case study adheres rigorously to the framework’s stages: conducting a comfort survey, developing comfort models, scaling up urban comfort predictions, and performing in-field validation.

    \begin{figure}[ht]
        \centering
        \includegraphics[width=0.9\linewidth]{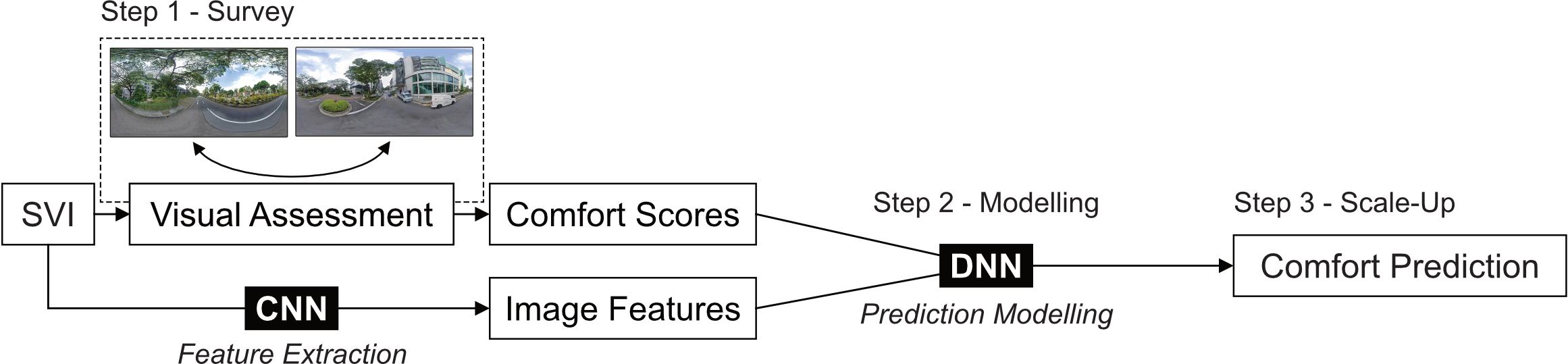}
        \caption{SVI-supported visual assessment as a replacement for in-field comfort survey.}
        \textbf{\label{fig:framework_case study_ai}}
    \end{figure}

In the comfort survey phase, we opted for a more efficient online approach using street view images instead of the traditional in-field survey combined with in-field validation, as shown in \textbf{Figure} \ref{fig:framework_case study_ai}. This method leverages information embedded in street view images and utilises AI-based CNN feature extraction to replace in-field surveys that typically require extensive environmental measurements, participant involvement, and significant human and financial resources. Nonetheless, for model validation, we still conducted a field comfort survey to confirm the model’s reliability.

\begin{figure}[ht]
        \centering
        \includegraphics[width=0.9\linewidth]{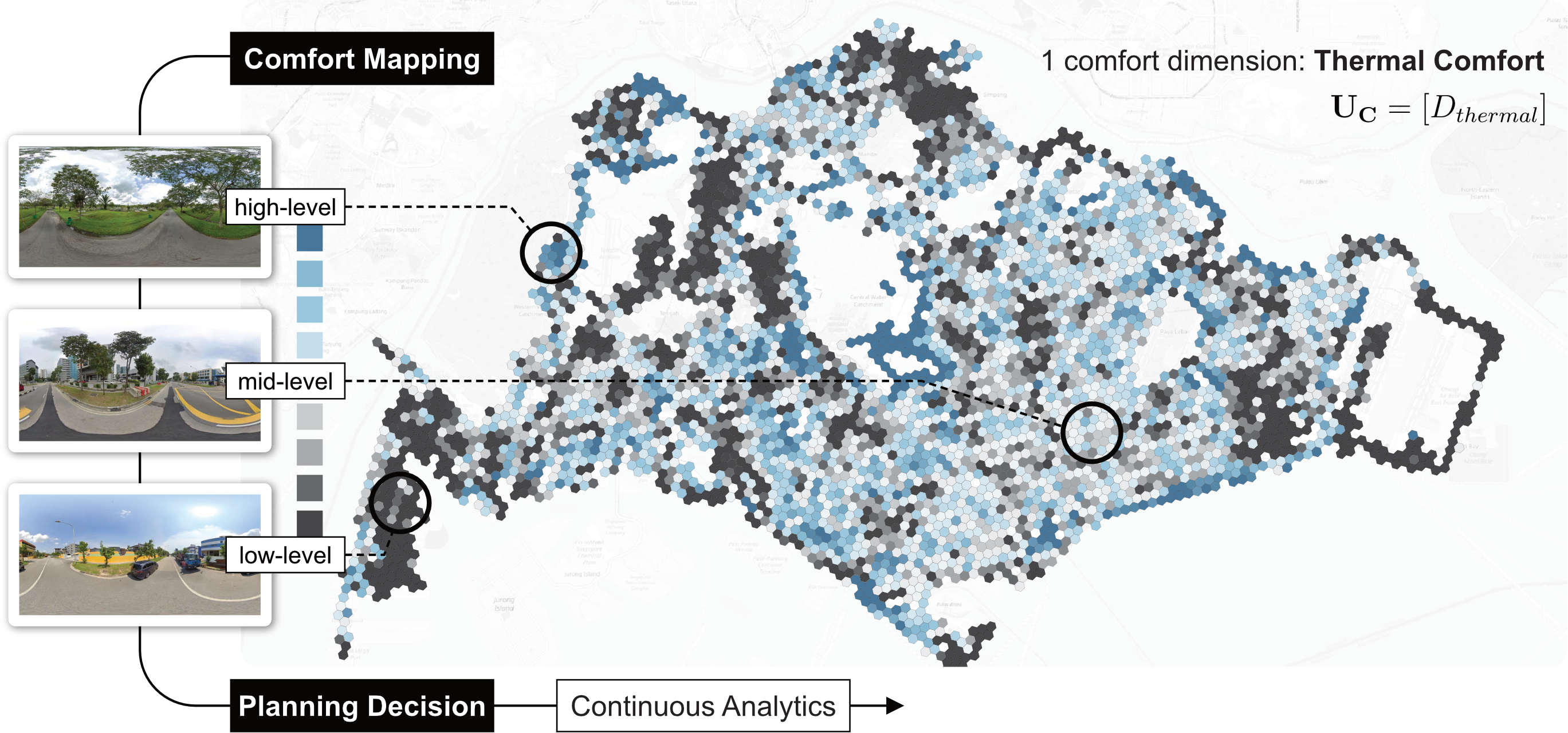}
        \caption{Comfort mapping for thermal comfort analytics in Singapore.}
        \textbf{\label{fig:case study_mapping}}
    \end{figure}

For comfort modelling, we developed two types of models: a comfort prediction model and a comfort inference model (\textbf{Figure} \ref{fig:framework_case study}). The prediction model employs a DNN to achieve high accuracy and generalisation capabilities, enabling us to apply it to 92,233 street view images across Singapore to predict thermal comfort levels along urban streets. The inference model, based on linear regression, interprets feature weights, allowing us to identify which factors significantly influence urban thermal comfort. Finally, we mapped Singapore’s urban thermal comfort (\textbf{Figure} \ref{fig:case study_mapping}), clearly identifying spatial areas with higher and lower comfort levels. As an addition, this mapping can provide valuable insights for urban planners to pinpoint zones requiring efforts to enhance street-level thermal comfort.

\section{Contributions and Limitations}

This paper establishes a multidimensional framework for understanding urban comfort, introducing four key steps for assessing urban comfort. Additionally, it examines the supportive role of AI in urban comfort modelling within the context of digital urban planning. In the case study, we apply this framework to assess urban-scale thermal comfort in Singapore, utilising CNN, DNN, and other advanced technologies for thermal comfort modelling. The resulting urban thermal comfort map of Singapore represents a valuable resource for digital urban planning efforts.

However, this study has several limitations. First, considering the notion of data sources, we incorporate three key modelling methods in this framework to support the objective. However, this approach does not fully account for the diverse auxiliary roles that various AI technologies could play in urban comfort modelling, depending on specific use cases. Future research could explore how different AI techniques—such as reinforcement learning, generative models, or multi-agent systems—can further enhance urban comfort assessment and digital planning. Second, the case study focuses solely on thermal comfort as a one-dimensional aspect of urban comfort, without extending to the multidimensional comfort model proposed in the framework. Future studies should consider developing and validating a multidimensional comfort model to substantiate the reliability of this conceptual framework for urban comfort assessment.

\section*{Acknowledgements}\label{acknowledgements}

We thank our colleagues at the NUS Urban Analytics Lab for the discussions. This research is supported by NUS Research Scholarship (NUSGS-CDE DO IS AY22\&L GRSUR0600042).
This research is part of the projects
(1) Multi-scale Digital Twins for the Urban Environment: From Heartbeats to Cities, which is supported by the Singapore Ministry of Education Academic Research Fund Tier 1;
(2) Large-scale 3D Geospatial Data for Urban Analytics, which is supported by the National University of Singapore under the Start Up Grant R-295-000-171-133;
(3) From Models to Pavements: Advancing Urban Digital Twins with a Multi-Dimensional Human-Centric Approach for a Smart Walkability Analysis, which is supported by the Humanities Social Sciences Seed Fund at the National University of Singapore (A-8001957-00-00).

\bibliographystyle{apalike}
\bibliography{references}

\begin{thebibliography}{}

\bibitem[Ayachi et~al., 2015]{ayachi2015identifying}
Ayachi, F., Dorey, J., and Guastavino, C. (2015).
\newblock Identifying factors of bicycle comfort: An online survey with
  enthusiast cyclists.
\newblock {\em Applied ergonomics}, 46:124--136.

\bibitem[Castaldo et~al., 2018]{castaldo2018subjective}
Castaldo, V.~L., Pigliautile, I., Rosso, F., Cotana, F., De~Giorgio, F., and
  Pisello, A.~L. (2018).
\newblock How subjective and non-physical parameters affect occupants’
  environmental comfort perception.
\newblock {\em Energy and Buildings}, 178:107--129.

\bibitem[Chwalek et~al., 2024]{chwalek2024dataset}
Chwalek, P., Zhong, S., Perry, N., Liu, T., Miller, C., Alavi, H.~S., Lalanne,
  D., and Paradiso, J.~A. (2024).
\newblock A dataset exploring urban comfort through novel wearables and
  environmental surveys.
\newblock {\em Scientific data}, 11(1):1423.

\bibitem[Du et~al., 2023]{du2023multiple}
Du, M., Hong, B., Gu, C., Li, Y., and Wang, Y. (2023).
\newblock Multiple effects of visual-acoustic-thermal perceptions on the
  overall comfort of elderly adults in residential outdoor environments.
\newblock {\em Energy and Buildings}, 283:112813.

\bibitem[Dubey et~al., 2016]{dubey2016deep}
Dubey, A., Naik, N., Parikh, D., Raskar, R., and Hidalgo, C.~A. (2016).
\newblock Deep learning the city: Quantifying urban perception at a global
  scale.
\newblock In {\em Computer Vision--ECCV 2016: 14th European Conference,
  Amsterdam, The Netherlands, October 11--14, 2016, Proceedings, Part I 14},
  pages 196--212. Springer.

\bibitem[Frontczak et~al., 2012]{frontczak2012quantitative}
Frontczak, M., Schiavon, S., Goins, J., Arens, E., Zhang, H., and Wargocki, P.
  (2012).
\newblock Quantitative relationships between occupant satisfaction and
  satisfaction aspects of indoor environmental quality and building design.
\newblock {\em Indoor air}, 22(2):119--131.

\bibitem[Guly{\'a}s et~al., 2006]{gulyas2006assessment}
Guly{\'a}s, {\'A}., Unger, J., and Matzarakis, A. (2006).
\newblock Assessment of the microclimatic and human comfort conditions in a
  complex urban environment: modelling and measurements.
\newblock {\em Building and Environment}, 41(12):1713--1722.

\bibitem[Ito et~al., 2024]{ito2024understanding}
Ito, K., Kang, Y., Zhang, Y., Zhang, F., and Biljecki, F. (2024).
\newblock Understanding urban perception with visual data: A systematic review.
\newblock {\em Cities}, 152:105169.

\bibitem[Kim et~al., 2022]{kim2022linking}
Kim, Y., Yu, S., Li, D., Gatson, S.~N., and Brown, R.~D. (2022).
\newblock Linking landscape spatial heterogeneity to urban heat island and
  outdoor human thermal comfort in tokyo: Application of the outdoor thermal
  comfort index.
\newblock {\em Sustainable Cities and Society}, 87:104262.

\bibitem[Klemm et~al., 2015]{klemm2015street}
Klemm, W., Heusinkveld, B.~G., Lenzholzer, S., and van Hove, B. (2015).
\newblock Street greenery and its physical and psychological impact on thermal
  comfort.
\newblock {\em Landscape and urban planning}, 138:87--98.

\bibitem[Lam et~al., 2020]{lam2020cross}
Lam, C. K.~C., Yang, H., Yang, X., Liu, J., Ou, C., Cui, S., Kong, X., and
  Hang, J. (2020).
\newblock Cross-modal effects of thermal and visual conditions on outdoor
  thermal and visual comfort perception.
\newblock {\em Building and Environment}, 186:107297.

\bibitem[Lei et~al., 2025]{lei2025developing}
Lei, B., Liu, P., Liang, X., Yan, Y., and Biljecki, F. (2025).
\newblock Developing the urban comfort index: Advancing liveability analytics
  with a multidimensional approach and explainable artificial intelligence.
\newblock {\em Sustainable Cities and Society}, page 106121.

\bibitem[Liang et~al., 2023]{liang2023revealing}
Liang, X., Zhao, T., and Biljecki, F. (2023).
\newblock Revealing spatio-temporal evolution of urban visual environments with
  street view imagery.
\newblock {\em Landscape and Urban Planning}, 237:104802.

\bibitem[Liu et~al., 2023]{liu2023towards}
Liu, P., Zhao, T., Luo, J., Lei, B., Frei, M., Miller, C., and Biljecki, F.
  (2023).
\newblock Towards human-centric digital twins: leveraging computer vision and
  graph models to predict outdoor comfort.
\newblock {\em Sustainable Cities and Society}, 93:104480.

\bibitem[Ma et~al., 2021]{ma2021critical}
Ma, X., Chau, C.~K., and Lai, J. H.~K. (2021).
\newblock Critical factors influencing the comfort evaluation for recreational
  walking in urban street environments.
\newblock {\em Cities}, 116:103286.

\bibitem[Miller et~al., 2025]{miller2025make}
Miller, C., Chua, Y.~X., Quintana, M., Lei, B., Biljecki, F., and Frei, M.
  (2025).
\newblock Make yourself comfortable: Nudging urban heat and noise mitigation
  with smartwatch-based just-in-time adaptive interventions (jitai).
\newblock {\em arXiv preprint arXiv:2501.09530}.

\bibitem[Miller et~al., 2023]{miller2023introducing}
Miller, C., Quintana, M., Frei, M., Chua, Y.~X., Fu, C., Picchetti, B., Yap,
  W., Chong, A., and Biljecki, F. (2023).
\newblock Introducing the cool, quiet city competition: Predicting
  smartwatch-reported heat and noise with digital twin metrics.
\newblock In {\em Proceedings of the 10th ACM International Conference on
  Systems for Energy-Efficient Buildings, Cities, and Transportation}, pages
  298--299.

\bibitem[Mosteiro-Romero et~al., 2024]{mosteiro2024people}
Mosteiro-Romero, M., Park, Y., and Miller, C. (2024).
\newblock People in cities: Combining subjective occupant feedback with
  urban-scale data to support indoor and outdoor thermal comfort.
\newblock In {\em The 5th Asia Conference of International Building Performance
  Simulation Association 2024}.

\bibitem[Qiu et~al., 2023]{qiu2023subjective}
Qiu, W., Li, W., Liu, X., Zhang, Z., Li, X., and Huang, X. (2023).
\newblock Subjective and objective measures of streetscape perceptions:
  Relationships with property value in shanghai.
\newblock {\em Cities}, 132:104037.

\bibitem[Rupp et~al., 2015]{rupp2015review}
Rupp, R.~F., V{\'a}squez, N.~G., and Lamberts, R. (2015).
\newblock A review of human thermal comfort in the built environment.
\newblock {\em Energy and buildings}, 105:178--205.

\bibitem[Salesses et~al., 2013]{salesses2013collaborative}
Salesses, P., Schechtner, K., and Hidalgo, C.~A. (2013).
\newblock The collaborative image of the city: mapping the inequality of urban
  perception.
\newblock {\em PloS one}, 8(7):e68400.

\bibitem[Santos~Nouri et~al., 2018]{santos2018approaches}
Santos~Nouri, A., Costa, J.~P., Santamouris, M., and Matzarakis, A. (2018).
\newblock Approaches to outdoor thermal comfort thresholds through public space
  design: A review.
\newblock {\em Atmosphere}, 9(3):108.

\bibitem[Shin, 2016]{shin2016toward}
Shin, J.-h. (2016).
\newblock Toward a theory of environmental satisfaction and human comfort: A
  process-oriented and contextually sensitive theoretical framework.
\newblock {\em Journal of Environmental Psychology}, 45:11--21.

\bibitem[Wang et~al., 2018]{wang2018uncertainty}
Wang, J., Wang, Z., de~Dear, R., Luo, M., Ghahramani, A., and Lin, B. (2018).
\newblock The uncertainty of subjective thermal comfort measurement.
\newblock {\em Energy and Buildings}, 181:38--49.

\bibitem[Xie et~al., 2019]{xie2019outdoor}
Xie, Y., Liu, J., Huang, T., Li, J., Niu, J., Mak, C.~M., and Lee, T.-c.
  (2019).
\newblock Outdoor thermal sensation and logistic regression analysis of comfort
  range of meteorological parameters in hong kong.
\newblock {\em Building and Environment}, 155:175--186.

\bibitem[Yang et~al., 2025]{yang2025thermal}
Yang, S., Chong, A., Liu, P., and Biljecki, F. (2025).
\newblock Thermal comfort in sight: Thermal affordance and its visual
  assessment for sustainable streetscape design.
\newblock {\em Building and Environment}, page 112569.

\bibitem[Yang et~al., 2023]{yang2023role}
Yang, S., Krenz, K., Qiu, W., and Li, W. (2023).
\newblock The role of subjective perceptions and objective measurements of the
  urban environment in explaining house prices in greater london: A multi-scale
  urban morphology analysis.
\newblock {\em ISPRS International Journal of Geo-Information}, 12(6):249.

\bibitem[Yang and Kang, 2005]{yang2005acoustic}
Yang, W. and Kang, J. (2005).
\newblock Acoustic comfort evaluation in urban open public spaces.
\newblock {\em Applied acoustics}, 66(2):211--229.

\bibitem[Yap et~al., 2023]{yap2023incorporating}
Yap, W., Chang, J.-H., and Biljecki, F. (2023).
\newblock Incorporating networks in semantic understanding of streetscapes:
  Contextualising active mobility decisions.
\newblock {\em Environment and Planning B: Urban Analytics and City Science},
  50(6):1416--1437.

\end{thebibliography}

\textbf{Biographies}\label{biography}

Sijie Yang is a PhD researcher at the NUS Urban Analytics Lab. He holds a MSc in Space Syntax from University College London. He is also a computer science master’s student at the School of Engineering and Applied Science, University of Pennsylvania.

Binyu Lei is a PhD researcher at Urban Analytics Lab, Department of Architecture in the College of Design and Engineering at National University of Singapore. She holds a Master degree in Urban Planning from the University of Melbourne.

Filip Biljecki is an assistant professor at the National University of Singapore and the founder of the NUS Urban Analytics Lab. He holds a MSc and PhD degree from the Delft University of Technology in the Netherlands. 

\end{document}